\documentclass[12pt]{amsart}
\usepackage[margin=1.1in]{geometry}
\usepackage{amssymb, amsmath, amscd, amsthm}
\usepackage{amsfonts}
\usepackage{psfrag}
\usepackage[all]{xy}
\usepackage{graphicx}
\usepackage{verbatim}
\usepackage{color}
\usepackage{tikz-cd}
\usepackage{todonotes}

\numberwithin{equation}{section}

\newtheorem{thm}{Theorem}[section]
\newtheorem{cor}[thm]{Corollary}
\newtheorem{lem}[thm]{Lemma}
\newtheorem{prop}[thm]{Proposition}
\theoremstyle{definition}
\newtheorem{defn}[thm]{Definition}
\theoremstyle{remark}
\newtheorem{rem}[thm]{Remark}
\numberwithin{equation}{subsection}
\newtheorem{ex}[thm]{Example}

\newcommand{\R}{\mathbb{R}}

\newcommand{\Aut}{\mathrm{Aut}}

\newcommand{\p}{\varphi}

\newcommand{\eps}{{\varepsilon}}

\usepackage{url,hyperref,lineno,microtype,subcaption}
\usepackage[onehalfspacing]{setspace}

\title[A topological model for partial equivariance]{A topological model for partial equivariance in deep learning and data analysis}
\author[Ferrari]{Lucia Ferrari}
\email{lucia.ferrari3@studio.unibo.it}
\address{}
\author[Frosini]{Patrizio Frosini }
\email{patrizio.frosini@unibo.it}
\address{}
\author[Quercioli]{Nicola Quercioli}
\email{nicola.quercioli2@unibo.it}
\address{}
\author[Tombari]{Francesca Tombari}
\email{tombari@kth.se}
\address{}


\begin{document}
\maketitle

\begin{abstract}
In this article, we propose a topological model to encode partial equivariance in neural networks. 
To this end, we introduce a class of operators, called P-GENEOs, that change data expressed by measurements, respecting the action of certain sets of transformations, in a non-expansive way. 
If the set of transformations acting is a group, then we obtain the so-called GENEOs.
We then study the spaces of measurements, whose domains are subject to the action of certain self-maps, and the space of P-GENEOs between these spaces. 
We define pseudo-metrics on them and show some properties of the resulting spaces.
In particular, we show how such spaces have convenient approximation and convexity properties.
\end{abstract}

\maketitle

\section{Introduction}

Over the past decade, several geometric techniques have been incorporated into Deep Learning (DL), giving rise to the new field of Geometric Deep Learning (GDL) (\cite{cohen2016group,masci2016geometric,7974879}.
This geometric approach to deep learning is exploited with a dual purpose. 
On one hand, geometry provides a common mathematical framework to study neural network architectures.
On the other hand, a geometric bias, based on prior knowledge of the data set, can be incorporated into DL models.
In this second case, GDL models take advantage of the symmetries imposed by an observer, which encode and elaborate the data. 
The general blueprint of many deep learning architectures is modelled by group equivariance to encode such properties.
If we consider measurements on a data set and a group encoding their symmetries, i.e., transformations taking admissible measurements to admissible measurements (for example, rotation or translation of an image), the group equivariance is the property guaranteeing that such symmetries are preserved after applying an operator (e.g., a layer in a neural network) on the observed data. 
In particular, taking the input measurements $\Phi$, the output measurements $\Psi$ and, respectively, their symmetry groups $G$ and $H$, the agent $F \colon \Phi \to \Psi$ is $T$-equivariant if $F(\varphi g) = F(\varphi) T(g)$, for any $\varphi$ in $\Phi$ and any $g$ in $G$, where $T$ is a group homomorphism from $G$ to $H$.
In the theory of Group Equivariant Non-Expansive Operators (GENEOs) (\cite{bergomi2019towards,bocchi2022geneonet,Bocchi2023,camporesi2018new,article2,cascarano2021geometric,Mi23,entropy-2410221}), as in many other GDL models, the collection of all symmetries is represented by a group, but in some applications, the group axioms do not necessarily hold since real-world data rarely follow strict mathematical symmetries due to noise, incompleteness or symmetry-breaking features. 
As an example, we can consider a data set that contains images of digits and the group of rotations as the group acting on it. 
Rotating an image of the digit `6' by a straight angle returns an image that the user would most likely interpret as `9'. 
At the same time, we may want to be able to rotate the digit `6' by small angles while preserving its meaning.

\begin{figure}[htp]
    \centering
    \includegraphics[width=11cm]{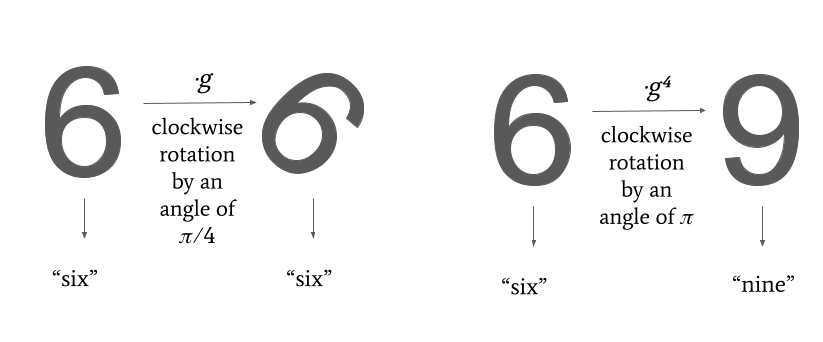}
    \caption{Example of a symmetry breaking feature. Applying a rotation $g$ of $\pi/4$, the digit `6' preserves its meaning (left). The rotation $g^4$ of $\pi$ is, instead, not admissible, since it transforms the digit `6' into the digit `9' (right).}
    \label{fig:six nine 01}
\end{figure}

It is then desirable to extend the theory of GENEOs by relaxing the hypotheses on sets of transformations. 
The main aim of this article is to give a generalization of the results obtained for GENEOs to a new mathematical framework where the property of equivariance is maintained only for some transformations of the measurements, encoding a partial equivariance with respect to the action of the group of all transformations. To this end, we introduce the concept of Partial Group Equivariant Non-Expansive Operator (P-GENEO).

In this new model there are some substantial differences with respect to the theory of GENEOs:
\begin{enumerate}
    \item The user chooses two sets of measurements in input: the one containing the original measurements and another set that encloses the admissible variations of such measurements, defined in the same domain. 
    For example, in the case where the function that represents the digit `6' is being observed, we define an initial space that contains this function and another space that contains certain small rotations of `6',  but excludes all the others. 
    \item Instead of considering a group of transformations we consider a set containing only those that do not change the meaning of our data, i.e., only those associating to each original measurement another one inside the set of its admissible variations.
    Therefore, by choosing the initial spaces, the user defines also which transformations of the data set, given by right composition, are admissible and which ones are not.
    \item We define partial GENEOs, or P-GENEOs, as a generalisation of GENEOs.
    P-GENEOs are operators that respect the two sets of measurements in input and the set of transformations relating them.
    The term partial refers to the fact that the set of transformations does not necessarily need to be a group.
\end{enumerate}
With these assumptions in mind we will extend the results proven in \cite{bergomi2019towards} and \cite{amsdottorato9770} for GENEOs. 
We will define suitable pseudo-metrics on the spaces of measurements, the set of transformations and the set of non-expansive operators. 
Grounding on their induced topological structures, we  prove compactness and convexity of the space of P-GENEOs, under the assumption that the function spaces are compact and convex.
These are useful properties from a computational point of view. 
For example, compactness guarantees that the space can be approximated by a finite set. 
Moreover, convexity allows us to take the convex combination of P-GENEOs in order to generate new ones.

 \section{Related work}
The main motivation for our work is that observed data rarely follow strict mathematical symmetries. 
This may be due, for example, to the presence of noise in data measurements. 
The idea of relaxing the hypothesis of equivariance in GDL and data analysis is not novel, as it is shown by the recent increase in the number of publications in this area 
(see, for example, \cite{RoLo22},\cite{van2022relaxing}, \cite{WaWaYu22}, \cite{NEURIPS2019_45d6637b}, \cite{NEURIPS2021_fc394e99} and \cite{nirvana}).

We identify two main ways to transform data via operators that are not strictly equivariant, due to the lack of strict symmetries of the measurements.
On one hand, one could define \textit{approximately equivariant} operator. 
These are operators for which equivariance holds up to small perturbation. 
In this case, given two groups, $G$ and $H$, acting on the spaces of measurements $\Phi$ and $\Psi$, respectively, and a homomorphism between them, $T\colon G\to H$, we say that $F\colon \Phi\to \Psi$ is $\varepsilon$-equivariant if, for any $g\in G$ and for any $\p\in \Phi$, $\|F(\p g)-F(\p)T(g)\|_\infty\le\eps$.
Alternatively, when defining operators transforming the measurements of certain data sets, equivariance may be substituted by \emph{partial equivariance}. 
In this case, equivariance is guaranteed for a subset of the groups acting on the space of measurements, with no guarantees for this subset to be a subgroup. 
Among the previously cited articles about relaxing the property of equivariance in DL, 
the approach of \cite{NEURIPS2021_fc394e99} is closer to an approximate equivariance model. 
There the authors use a Bayesian approach to introduce an inductive bias in their network that is sensitive to approximate symmetry. 
The authors of \cite{RoLo22}, instead, utilize a partial equivariance approach, where a probability distribution is defined and associated with each group convolutional layer of the architecture and, the parameter defining it are either learnt, to achieve equivariance, or partially learnt, to achieve partial equivariance. 
The importance of choosing equivariance with respect to different acting groups on each layer of the CNN was actually first observed in \cite{NEURIPS2019_45d6637b} for the group of Euclidean isometries in $\R^2$.

The point of view of this article is closer to the latter. 
Our P-GENEOs are indeed operators that preserve the action of certain sets ruling the admissibility of the transformations of the measurements of our data sets. 
Moreover, non-expansiveness plays a crucial role in our model.
This is, in fact, the feature allowing us to obtain compactness and approximability in the space of operators, and distinguishing our model from the existing literature on equivariant machine learning.

\section{Mathematical setting}
\subsection{Data sets and operations}

Consider a set $X$ and the normed vector space $(\R_b^X, \lVert \cdot \rVert_\infty)$, where $\R_b^X$ is the space of all bounded real-valued functions on $X$ and $\lVert \cdot \rVert_\infty$ is the usual uniform norm, i.e., for any $f \in \R_b^X$, $\lVert f \rVert_\infty := \sup_{x \in X} \lvert f(x) \rvert$. On the set $X$ the space of transformations is given by elements of $\Aut(X)$, i.e., the group of bijections from $X$ to itself. Then, we can consider the right group action $\mathcal{R}$ defined as follows (we represent composition as a juxtaposition of functions):
\[
\mathcal{R}\colon \R_b^X\times \mathrm{Aut}(X)\to \R_b^X,\ \ \ \ (\varphi,s)\mapsto \varphi s.
\]

\begin{rem}
For every $s\in\mathrm{Aut}(X)$, the map $\mathcal{R}_s\colon\R_b^X\to \R_b^X,$ with $\mathcal{R}_s(\varphi) := \varphi s$ preserves the distances. In fact, for any $\p_1,\p_2\in\R_b^X$, by bijectivity of $s$, we have that
\begin{align*}
    \lVert \mathcal{R}_s(\p_1)-\mathcal{R}_s(\p_2)\rVert_\infty&=\sup_{x\in X}|\p_1  s(x)-\p_2  s(x)|\\
    &= \sup_{y\in X}|\p_1(y)-\p_2(y)|\\
    &=\lVert \p_1-\p_2\rVert_\infty.
\end{align*}
\end{rem}

In our model our data sets are represented as two sets $\Phi$ and $\Phi'$ of bounded real-valued measurements on $X$.
In particular, $X$ represents the space where the measurements can be made, $\Phi$ is the space of permissible measurements,  and ${\Phi'}$ is a space which $\Phi$ can be transformed into, without changing the interpretation of its measurements after a transformation is applied.
In other words, we want to be able to apply some admissible transformations on the space $X$, so that the resulting changes in the measurements in $\Phi$ are contained in the space $\Phi'$. Thus, in our model, we consider operations on $X$ in the following way:
\begin{defn}\label{defn_aut}
A {\bf $(\Phi,{\Phi'})$-operation} is an element $s$ of $\Aut(X)$ such that, for any measurement $\varphi \in \Phi$, the composition $\varphi s$ belongs to ${\Phi'}$.
The set of all $(\Phi,{\Phi'})$-operations is denoted by $\text{Aut}_{\Phi, {\Phi'}}(X)$.
\end{defn}
\begin{rem} \label{rem_id}
    We can observe that the identity function $\text{id}_X$ is an element of $\Aut_{\Phi,\Phi'}(X)$ if and only if $\Phi\subseteq\Phi'$.
\end{rem}
For any $s \in \Aut_{\Phi, {\Phi'}}(X)$, the restriction to $\Phi \times \mathrm{Aut}_{\Phi,{\Phi'}}(X)$ of the map $\mathcal{R}_s$ takes values in $\Phi'$ since $\mathcal{R}_s(\varphi):=\varphi s\in \Phi'$ for any $\p\in\Phi$.
We can consider the restriction of the map $\mathcal{R}$ (for simplicity, we will continue to use the same symbol to denote this restriction):
\[
\mathcal{R}\colon\Phi\times \mathrm{Aut}_{\Phi,{\Phi'}}(X)\to {\Phi'},\ \ \ \ (\varphi,s)\mapsto \varphi s
\]
where $\mathcal{R}(\p,s)=\mathcal{R}_s(\p)$, for every $s\in\Aut_{\Phi,{\Phi'}}(X)$ and every $\p\in\Phi$.





\begin{defn}
    Let $X$ be a set. A {\bf perception triple} is a triple $(\Phi,{\Phi'},S)$ with $\Phi, {\Phi'} \subseteq \R_b^X$ and $S\subseteq\Aut_{\Phi,{\Phi'}}(X)$. The set $X$ is called the \textbf{domain} of the perception triple and is denoted by $\mathrm{dom}(\Phi,{\Phi'},S)$.
\end{defn}

\begin{ex}
Given $X=\R^2$, consider two rectangles $R$ and $R'$ in $X$. Assume $\Phi:=\{\p\colon X\to [0,1] : \ \text{supp}(\p)\subseteq R\}$ and ${\Phi'}:=\{{\p'}\colon X\to [0,1] : \ \text{supp}({\p'})\subseteq R'\}$.  We recall that, if we consider a function $f\colon X\to \R$, the \textit{support} of $f$ is the set of points in the domain where the function does not vanish, i.e., $\text{supp}(f)=\{x\in X \ | \ f(x)\neq 0$\}.
Consider $S$ as the set of translations that bring $R$ into $R'$. 
The triple $(\Phi,{\Phi'},S)$ is a perception triple.
If $\Phi$ represents a set of grey level images, $S$ determines which translations can be applied to our pictures.
\end{ex}

\subsection{Pseudo-metrics on data sets}
In our model, data are represented as function spaces, that is, considering a generic set $X$, sets $\Omega\subseteq\R_b^X$ of bounded real-valued functions. We endow the real line $\R$ with the usual Euclidean metric and the space $X$ with an extended pseudo-metric induced by $\Omega$:
\[D_{X}^{\Omega}(x_1,x_2)=\sup_{\omega \in \Omega} \lvert \omega(x_1)-\omega(x_2) \rvert\]
for every $x_1, x_2\in X$. 
The choice of this pseudo-metric over X means that two points can only be distinguished if they assume different values for some measurements. For example, if $\Phi$ contains only a constant function and $X$ contains at least two points, the distance between any two points of $X$ is always null.

The pseudo-metric space $X_\Omega:=(X,D_X^\Omega)$ can be considered as a topological space with the basis 
\[
\mathcal{B}_\Omega=\{B_\Omega(x_0,r)\}_{x_0\in X, \ r\in \R^+}=\big\{\{x \in X : \ D_{X}^{\Omega}(x,x_0)<r\}\big\}_{x_0\in X, \ r\in \R^+},
\]
and the induced topology is denoted by $\tau_\Omega$. The reason for considering a topological space $X$, rather than just a set, follows from the need of formalising the assumption that data are stable under small perturbations.
\begin{rem} \label{rem_pseudo-metrics}
    In our case, there are two collections of functions $\Phi$ and ${\Phi'}$ in $\R_b^X$ representing our data, both of which induce a topology on $X$.
    Hence, in the model, we consider two pseudo-metric spaces $X_\Phi$ and $X_{\Phi'}$ with the same underlying set $X$.
    If $\Phi\subseteq\Phi'\subseteq\R_b^X$, then the topologies $\tau_{\Phi'}$ and $\tau_{\Phi'}$ are comparable and, in particular, $\tau_{\Phi'}$ is finer than $\tau_\Phi$.
\end{rem}


Now, given a set $\Omega\subseteq\R_b^X$, we will prove a result about the compactness of the pseudo-metric space $X_\Omega$. Before proceeding, let us
recall the following lemma (e.g., see \cite{gaal1964point}):
\begin{lem} \label{lem_totbound}
    Let (P,d) be a pseudo-metric space. The following conditions are equivalent:
    \begin{enumerate}
        \item P is totally bounded;
        \item every sequence in P admits a Cauchy subsequence.
    \end{enumerate}
\end{lem}
    

\begin{thm} \label{thm_X_totbound}
    If $\Omega$ is totally bounded, then $X_\Omega$ is totally bounded. 
\end{thm}
\begin{proof}
    By Lemma \ref{lem_totbound} it will suffice to prove that every sequence in $X$ admits a Cauchy subsequence with respect to the pseudo-metric $D_X^\Omega$. 
    Consider a sequence $(x_i)_{i\in\mathbb{N}}$ in $X_\Omega$ and take a real number $\eps>0$. Since $\Omega$ is totally bounded, we can find a finite subset $\Omega_\eps=\{\omega_1,\dots,\omega_n\}$ such that for every $\omega\in\Omega$ there exists $\omega_r\in\Omega$ for which $\|\omega-\omega_r\|_\infty<\eps$. 
    We can consider now the real sequence $(\omega_1(x_i))_{i\in\mathbb{N}}$, which is bounded since $\Omega\in\R_b^X$.
    From Bolzano-Weierstrass Theorem it follows that we can extract a convergent subsequence $(\omega_1(x_{i_h}))_{h\in\mathbb{N}}$. 
    Again, we can extract from $(\omega_2(x_{i_h}))_{h\in\mathbb{N}}$ another convergent subsequence $(\omega_2(x_{i_{h_t}}))_{t\in\mathbb{N}}$.
    Repeating the process, we are able to extract a subsequence of $(x_i)_{i\in\mathbb{N}}$, that for simplicity of notation we can indicate as $(x_{i_j})_{j\in\mathbb{N}}$, such that $(\omega_k(x_{i_j}))_{j\in\mathbb{N}}$ is a convergent subsequence in $\R$, and hence a Cauchy sequence in $\R$, for every $k\in\{1,\dots,n\}$.
    By construction, $\Omega_\eps$ is finite, then we can find an index $\bar{\jmath}$ such that for any $k\in\{1,\dots, n\}$
    \begin{equation*}
        |\omega_k(x_{i_\ell})-\omega_k(x_{i_m})|\le\eps,\quad \text{for every} \quad \ell,m\geq\bar{\jmath}.
    \end{equation*}
    Furthermore we have that, for any $\omega \in \Omega$, any $\omega_k\in\Omega_\eps$ and any $\ell,m\in\mathbb{N}$
    \begin{align*}
         |\omega (x_{i_\ell})-\omega (x_{i_m})|&\le  |\omega (x_{i_\ell})-\omega_k (x_{i_\ell})|+ |\omega_k (x_{i_\ell})-\omega_k (x_{i_m})| +|\omega_k (x_{i_m})-\omega (x_{i_m})|\\
         &\le\|\omega-\omega_k\|_\infty+|\omega_k (x_{i_\ell})-\omega_k (x_{i_m})|+\|\omega_k-\omega\|_\infty.
    \end{align*}
    We observe that the choice of $\Bar{\jmath}$ depends only on $\eps$ and $\Omega_\eps$, not on $k$. Then, choosing a $\omega_k\in\Omega_\eps$ such that $\|\omega_k-\omega\|_\infty<\eps$, we get $\|\omega (x_{i_\ell})-\omega (x_{i_m})\|_\infty<3\eps$ for every $\omega \in \Omega$ and every $l,m\geq\bar{\jmath}$. Then,
    \begin{equation*}
        D_X^\Omega(x_{i_\ell}, x_{i_m})=\sup_{\omega\in\Omega}|\omega(x_{i_\ell})-\omega(x_{i_m})|<3\eps \quad \text{for every} \quad \ell,m\geq\bar{\jmath}.
    \end{equation*}
    Then $(x_{i_j})_{j\in\mathbb{N}}$ is a Cauchy sequence in $X_\Omega$. For Lemma \ref{lem_totbound} the statement holds.
\end{proof}

\begin{cor}
If $\Omega$ is totally bounded and $X_\Omega$ is complete, then $X_\Omega$ is compact.
\end{cor}
\begin{proof}
    From Theorem \ref{thm_X_totbound} we have that $X_\Omega$ is totally bounded and since by hypothesis it is also complete, it is compact. 
\end{proof}
 
Now, we will prove that the choice of the  pseudo-metric $D_X^\Omega$ on $X$ makes the functions in $\Omega$ non-expansive.
\begin{defn}
Consider two pseudo-metric spaces $(P,d_P)$ and $(Q, d_Q)$. A function $f\colon P\to Q$ is \textbf{non-expansive} 
if  $d_Q(f(p_1),f(p_2))\le d_P(p_1,p_2)$ for any $p_1,p_2\in P$.\\ 
We denote by $\mathbf{NE}(P,Q)$ the space of all non-expansive functions from $(P,d_P)$ to $(Q, d_Q)$.
\end{defn}

\begin{prop} \label{prop_NE}
    $\Omega \subseteq \mathbf{NE}(X_\Omega,\R)$.
\end{prop}
\begin{proof}
    For any $x_1,x_2 \in X$ we have that
    \[\lvert \omega(x_1)-\omega(x_2) \rvert \le \sup_{\omega \in \Omega} \lvert \omega(x_1)-\omega(x_2) \rvert=D_{X}^{\Omega}(x_1,x_2).\]
\end{proof}

Then, the topology on X induced by $D_X^\Omega$ naturally makes the measurements in $\Omega$ continuous.
In particular, since the previous results hold for a generic $\Omega \subseteq \R_b^X$, they are also true for $\Phi$ and ${\Phi'}$ in our model.

\begin{rem}
    Assume that $(\Phi,\Phi',S)$ is a perception triple. A function ${\p'}\in{\Phi'}$ may not be continuous from $X_\Phi$ to $\R$ and a function $\p\in\Phi$ may not be continuous from $X_{\Phi'}$ to $\R$. In other words, the topology on $X$ induced by the pseudo-metric of one of the function spaces does not make the functions in the other continuous.
\end{rem}
\begin{ex}
Assume $X=\R$ and for every $a,b\in\R$ consider the functions $\p_a\colon X\to \R$ and $\p'_b\colon X\to \R$ defined by setting 
    \begin{equation*}
    \p_a(x)=
    \begin{cases} 
        0 \quad &\text{if} \ x\geq a \\
        1 \quad &\text{otherwise}    
    \end{cases} 
    ,\quad \qquad \p'_b(x)=
    \begin{cases} 
        0 \quad &\text{if} \ x\le b \\
        1 \quad &\text{otherwise}    
    \end{cases}.
    \end{equation*}
    Suppose $\Phi:=\{\p_a:a\geq 0\}$ and $\Phi':=\{\p'_b:b\le 0\}$, and consider the symmetry with respect to the y-axis, i.e., the map $s(x)=-x$. Surely, $s\in \Aut_{\Phi,\Phi'}(X)$. We can observe that the function $\p_1\in\Phi$ is not continuous from $X_\Phi'$ to $\R$; indeed $D_X^{\Phi'}(0,2)=0$, but $|\p_1(0)-\p_1(2)|=1$.
\end{ex}
However, if $\Phi\subseteq\Phi'$, we have that the functions in $\Phi$ are also continuous on $X_{\Phi'}$, indeed:
\begin{cor}
 If $\Phi\subseteq\Phi'$, then ${\Phi}\subseteq\textbf{NE}(X_{\Phi'},\R)$.
\end{cor}
\begin{proof}
    By Proposition \ref{prop_NE} the statement trivially holds since $\Phi\subseteq\Phi'\subseteq\textbf{NE}(X_{\Phi'},\R)$.
\end{proof}

\subsection{Pseudo-metrics on the space of operations}
\begin{prop} \label{prop_s_ne}
Every element of $\Aut_{\Phi, {\Phi'}}(X)$ is non-expansive from $X_{\Phi'}$ to $X_{\Phi}$.
\end{prop}

\begin{proof}
   Considering a bijection $s\in \Aut_{\Phi, {\Phi'}}(X)$ we have that
    \begin{align*}
        D_X^\Phi(s(x_1),s(x_2))&=\sup_{\p\in\Phi}|\p s(x_1)-\p s(x_2)|\\
        &=\sup_{\p\in\Phi s} |\p(x_1)-\p(x_2)|\\
        &\le \sup_{{\p'}\in{\Phi'}}|{\p'} (x_1)-{\p'} (x_2)|=D_X^{{\Phi'}}(x_1,x_2)
    \end{align*}
    for every $x_1,x_2\in X$, where $\Phi s = \{ \varphi s, \varphi \in \Phi \}$.
    Then, $s\in \textbf{NE}(X_{{\Phi'}},X_\Phi)$ 
    and the statement is proved.
\end{proof}

Now we are ready to put more structure on $\Aut_{\Phi, {\Phi'}}(X)$. 
Considering a set $\Omega\subseteq \R_b^X$ of bounded real-valued functions, we can endow the set $\Aut(X)$ with a pseudo-metric inherited from $\Omega$:
\[
D_\Aut^\Omega(s_1,s_2):=\sup_{\omega \in \Omega}\lVert \omega s_1-\omega s_2\rVert_\infty
\]
for any $s_1,s_2$ in $\Aut(X)$.
\begin{rem}
    Analogously to what happens in Remark \ref{rem_pseudo-metrics} for $X$, the sets $\Phi$ and $\Phi'$ can endow $\Aut(X)$ with two possibly different pseudo-metrics $D_\Aut^\Phi$ and $D_\Aut^{\Phi'}$.
    In particular, we can consider $\Aut_{\Phi,{\Phi'}}(X)$ as a pseudo-metric subspace of $\Aut(X)$ with the induced pseudo-metrics.
\end{rem}


\begin{rem}
    We observe that, for any $s_1,s_2$ in $\Aut(X)$, 
\begin{align} \label{eq:digi2}
D_\Aut^\Omega(s_1,s_2)&:=\sup_{\omega \in \Omega}\lVert\omega s_1 - \omega s_2 \rVert_\infty\nonumber\\
&=\sup_{x \in X} \sup_{\omega \in \Omega} | \omega(s_1(x)) - \omega(s_2(x)) |\nonumber\\
&=\sup_{x \in X}D_X^{\Omega}(s_1(x),s_2(x)).
\end{align}
In other words, the pseudo-metric $D_\mathrm{Aut}^\Omega$, which is based on the action of the elements of $\Aut(X)$ on the set $\Omega$, is exactly the usual uniform pseudo-metric on $X_\Omega$.

\end{rem}
\subsection{The space of operations}
Since we are only interested in transformations of functions in $\Phi$, it would be natural to just endow $\mathrm{Aut}_{\Phi, \Phi’}(X)$ with the pseudo-metric $D_\mathrm{Aut}^\Phi$. However, it is sometimes necessary to consider the pseudo-metric $D_{\mathrm{Aut}}^{\Phi’}$ in order to guarantee the continuity of the composition of elements in $\mathrm{Aut}_{\Phi, \Phi’}(X)$, whenever it is admissible. Consider two elements $s,t$ in $\Aut_{\Phi,\Phi'}(X)$ such that $st$ is still an element of $\Aut_{\Phi,\Phi'}(X)$, i.e., for every function $\p\in\Phi$ we have that $\p st \in \Phi'$. Then, for any $\p\in\Phi$ we have that
\begin{equation*}
    \p':=\p s \in \Phi s \subseteq\Phi', \quad \p' t\in\Phi'.
\end{equation*}
Therefore, $t$ is also an element of $\Aut_{\Phi s,\Phi'}(X)$. By definition $\Phi s$ is contained in $\Phi'$ for every $s\in\Aut_{\Phi,\Phi'}(X)$ and this justifies the choice of considering in $\Aut_{\Phi,\Phi'}(X)$ also the pseudo-metric $D_\Aut^{\Phi'}$.
We have shown in particular that if $s,t$ are elements of $\Aut_{\Phi,\Phi'}(X)$ such that $st$ is still an element of $\Aut_{\Phi,\Phi'}(X)$, then $t$ is an element of $\Aut_{\Phi s,\Phi'}(X)$, which is an implication of the following proposition:
\begin{prop} \label{prop_composition}
Let $s,t\in\Aut_{\Phi,\Phi'}(X)$. Then $st\in\Aut_{\Phi,\Phi'}(X)$ if and only if $t\in\Aut_{\Phi s,\Phi'}(X)$.
\end{prop}
\begin{proof}
If the composition $st$ belongs to $\Aut_{\Phi,\Phi'}(X)$, we have already proved that $t\in\Aut_{\Phi s,\Phi'}(X)$.
On the other hand, if $t\in\Aut_{\Phi s,\Phi'}(X)$ we have that $\bar{\p} t\in\Phi'$ for every $\bar{\p}\in\Phi s$. 
Since $\p(st)=(\p s) t$, it follows that $\p(st)\in\Phi'$ for every $\p\in\Phi$. Therefore,  $st\in\Aut_{\Phi,\Phi'}(X)$ and the statement is proved.
\end{proof}

\begin{rem} \label{rem_comp}
Let $t\in\Aut_{\Phi,\Phi'}(X)$. We can observe that if $s\in\Aut_{\Phi}(X)$, then $\Phi s\subseteq\Phi$ and $st\in\Aut_{\Phi,\Phi'}(X)$.
\end{rem}

\begin{lem}\label{ipghjeruioh}
Consider $r,s,t \in \mathrm{Aut}(X)$. For any $\Omega \subseteq \R^X_b$, it holds that
\[
D_\Aut^\Omega (rt,st)=D_\Aut^\Omega (r,s).
\]
\end{lem}

\begin{proof}
	Since $\mathcal{R}_t$ preserves the distances, we have that:
	\begin{align*}
		D_\Aut^\Omega(rt,st)&:= \sup_{\omega \in \Omega}\| \omega rt - \omega st\|_\infty\\
		& = \sup_{\omega \in \Phi}\| \omega r - \omega s\|_\infty\\
		&= D_\Aut^\Omega(r,s).
	\end{align*}
\end{proof}
\begin{lem} \label{lem_comp_sx}
    Consider $r,s\in \Aut(X)$ and $t\in \Aut_{\Phi,\Phi'}(X)$. It holds that
    \[D_\Aut^\Phi(tr,ts)\le D_\Aut^{\Phi'}(r,s).\]
\end{lem}
\begin{proof}
 Since $\Phi t \subseteq \Phi'$, we have that:
\begin{align*}
    D_\Aut^\Phi(tr,ts)&=\sup_{\p\in\Phi}\|\p tr -\p ts\|_\infty\\
    &= \sup_{\p'\in\Phi t}\|\p'r -\p's\|_\infty\\
    &\le \sup_{\p'\in\Phi'}\|\p'r -\p's\|_\infty\\
    &= D_\Aut^{\Phi'}(r,s).
\end{align*}
    
\end{proof}

Let $\Pi$ be the set of all pairs $(s,t)$ such that $s,t,st\in\Aut_{\Phi, {\Phi'}}(X)$. We endow $\Pi$ with the pseudo-metric
    \[D_\Pi((s_1,t_1),(s_2,t_2)):=D_\Aut^\Phi(s_1,s_2)+D_\Aut^{\Phi'}(t_1,t_2)\]
and the corresponding topology.
\begin{prop} \label{prop_Pi}
The function $\circ \colon \Pi\to (\Aut_{\Phi, {\Phi'}}(X),D_\Aut^\Phi)$ that maps $(s,t)$ to $st$ is non-expansive, and hence continuous. 
\end{prop}

\begin{proof}
Consider two elements $(s_1,t_1), (s_2,t_2)$ of $\Pi$. By Lemma \ref{ipghjeruioh} and Lemma \ref{lem_comp_sx},
\begin{align*}
    D_\Aut^{\Phi}(s_1 t_1, s_2 t_2)&\le D_\Aut^{\Phi}(s_1 t_1, s_2 t_1) + D_\Aut^{\Phi}(s_2 t_1, s_2 t_2)\\
    &\le D_\Aut^{\Phi}(s_1, s_2) + D_\Aut^{\Phi'}(t_1, t_2)\\
    &= D_\Pi((s_1,t_1),(s_2,t_2)).
\end{align*}
Therefore, the statement is proved.
\end{proof}

Let $\Upsilon$ be the set of all $s$ with  $s,s^{-1}\in\Aut_{\Phi, {\Phi'}}(X)$.

\begin{prop}
The function $(\cdot)^{-1} \colon (\Upsilon,D_\Aut^{\Phi'}) \to (\Aut_{\Phi, {\Phi'}}(X),D_\Aut^\Phi)$, that maps $s$ to $s^{-1}$, is non-expansive, and hence continuous.
\end{prop}
\begin{proof}
    Consider two bijections $s_1,s_2\in\Upsilon$. Because of Lemma \ref{ipghjeruioh} and Lemma \ref{lem_comp_sx}, we obtain that
    \begin{align*}
        D_{\Aut}^{\Phi}(s_1^{-1},s_2^{-1})&=D_{\Aut}^{\Phi}(s_1^{-1}s_2,s_2^{-1}s_2)\\
        &=D_{\Aut}^{\Phi}(s_1^{-1}s_2,\mathrm{id}_X)\\
        &=D_{\Aut}^{\Phi}(s_1^{-1}s_2,s_1^{-1}s_1)\\
        &\le D_{\Aut}^{\Phi'}(s_2,s_1) = D_{\Aut}^{\Phi'}(s_1,s_2).
    \end{align*}
\end{proof}

We have previously defined the map
\[
\mathcal{R}\colon\Phi\times \mathrm{Aut}_{\Phi,{\Phi'}}(X)\to {\Phi'},\ \ \ \ (\varphi,s)\mapsto \varphi s
\]
where $\mathcal{R}(\Phi,s)=\mathcal{R}_s(\Phi)$, for every $s\in\Aut_{\Phi,{\Phi'}}(X)$.

\begin{prop} \label{azione_continua}
The function $\mathcal{R}$ is continuous, by choosing the pseudo-metric $D_\Aut^\Phi$ on $\Aut_{\Phi, {\Phi'}}(X)$.
\end{prop}
\begin{proof}
    We have that
	   \begin{align*}
		\lVert \mathcal{R}(\varphi,t) - \mathcal{R}(\overline{\p}, s) \rVert_\infty& = \lVert \varphi t - \overline{\p} s \rVert_\infty \\
		& \le \|\varphi t-\varphi s\|_\infty +\|\varphi s-\overline{\p} s\|_\infty\nonumber\\
		& = \|\varphi t-\varphi s\|_\infty +\|\varphi-\overline{\p}\|_\infty\nonumber\\
		& \le D_\Aut^\Phi(t,s)+\|\varphi-\overline{\p}\|_\infty 
		\end{align*}
  for any $\varphi, \overline{\p} \in \Phi$ and any $t,s \in \Aut_{\Phi,{\Phi'}}(X)$.
    This proves that $\mathcal{R}$ is continuous.
\end{proof}

Now,  we can give a result about the compactness of $(\Aut_{\Phi,\Phi'}(X),D_\Aut^\Phi)$, under suitable assumptions.
\begin{prop} \label{thm_S_totbound}
If $\Phi$ and ${\Phi'}$ are totally bounded, then $(\Aut_{\Phi,\Phi'}(X),D_\Aut^\Phi)$ is totally bounded.
\end{prop}
\begin{proof}
     Consider a sequence $(s_i)_{i\in\mathbb{N}}$ in $\Aut_{\Phi,\Phi'}(X)$ and a real number $\eps>0$. Since $\Phi$ is totally bounded, we can find a finite subset $\Phi_\eps=\{\p_1,\dots,\p_n\}$ such that for every $\p\in\Phi$ there exists $\p_r\in\Phi$ for which $\|\p-\p_r\|_\infty<\eps$. Now, consider the sequence $(\p_1 s_i)_{i\in\mathbb{N}}$ in ${\Phi'}$. Since also ${\Phi'}$ is totally bounded, from Lemma \ref{lem_totbound} it follows that we can extract a Cauchy subsequence $(\p_1 s_{i_h})_{h\in\mathbb{N}}$. Again, we can extract another Cauchy subsequence $(\p_2s_{i_{h_t}})_{t\in\mathbb{N}}$. Repeating the process for every $k\in\{1,\dots, n\}$, we are able to extract a subsequence of $(s_i)_{i\in\mathbb{N}}$, that for simplicity of notation we can indicate as $(s_{i_j})_{j\in\mathbb{N}}$, such that $(\p_k s_{i_j})_{j\in\mathbb{N}}$ is a Cauchy sequence in $\Phi'$ for every $k\in\{1,\dots, n\}$.\\
     By definition $\Phi_\eps$ is finite, then we can find an index $\bar{\jmath}$ such that for any $k\in\{1,\dots, n\}$
    \begin{equation} \label{eq_tot_bound}
        \|\p_k s_{i_\ell}-\p_k s_{i_m}\|_\infty\le\eps,\quad \text{for every} \quad \ell,m\geq\bar{\jmath}.
    \end{equation}
    Furthermore we have that, for any $\p \in \Phi$, any $\p_k\in\Phi_\eps$ and any $\ell,m\in\mathbb{N}$
    \begin{align*}
         \|\p s_{i_\ell}-\p s_{i_m}\|_\infty&\le  \|\p s_{i_\ell}-\p_k s_{i_\ell}\|_\infty+ \|\p_k s_{i_\ell}-\p_k s_{i_m}\|_\infty+ \|\p_k s_{i_m}-\p s_{i_m}\|_\infty\\
         &=\|\p-\p_k\|_\infty+\|\p_k s_{i_\ell}-\p_k s_{i_m}\|_\infty+\|\p_k-\p\|_\infty.
    \end{align*}
    We observe that the choice of $\Bar{\jmath}$ in \eqref{eq_tot_bound} depends only on $\eps$ and $\Phi_\eps$, not on $\varphi$. Then, choosing a $\p_k\in\Phi_\eps$ such that $\|\p_k-\p\|_\infty<\eps$, we get $\|\p s_{i_\ell}-\p s_{i_m}\|_\infty<3\eps$ for every $\p \in \Phi$ and every $\ell,m\geq\bar{\jmath}$. Hence, for every $\ell,m \in \mathbb{N}$
    \begin{equation*}
        D_\Aut^\Phi(s_{i_\ell}, s_{i_m})=\sup_{\p\in\Phi}\|\p s_{i_\ell}-\p s_{i_m}\|_\infty<3\eps
    \end{equation*}
    Therefore $(s_{i_j})_{j\in\mathbb{N}}$ is a Cauchy sequence in $\Aut_{\Phi,\Phi'}(X)$. For Lemma \ref{lem_totbound} the statement holds.
\end{proof}

\begin{cor}
    Assume that $S \subseteq \Aut_{\Phi,\Phi'}(X)$. If $\Phi$ and ${\Phi'}$ are totally bounded and $(S,D_\Aut^\Phi)$ is complete, then it is also compact.
\end{cor}
\begin{proof}
    From Proposition \ref{thm_S_totbound} we have that $S$ is totally bounded and since by hypothesis it is also complete, the statement holds.
\end{proof}

\section{The space of P-GENEOs} \label{chap2}
In this section we introduce the concept of Partial Group Equivariant Non-Expansive Operator (P-GENEO). P-GENEOs allow us to transform data sets, preserving symmetries and distances and maintaining the acceptability conditions of the transformations. We will also describe some topological results about the structure of the space of P-GENEOs and some techniques used for defining new P-GENEOs in order to populate the space of P-GENEOs.
\begin{defn}\label{wfbjbfrbhjf}
    Let $X,Y$ be sets and $(\Phi,{\Phi'},S)$, $(\Psi,{\Psi'},Q)$ be perception triples with domains $X$ and $Y$, respectively. 
Consider a triple of functions $(F,F',T)$ with the following properties: 
\begin{itemize}
    \item $F \colon \Phi \to \Psi$, $F' \colon {\Phi'} \to \Psi'$, $T \colon S\to Q$;
    \item for any $s,t \in S$ such that $st \in S$ it holds that $T(st)=T(s)T(t);$
    \item for any $s\in S$ such that $s^{-1}\in S$ it holds that $T(s^{-1})=T(s)^{-1}$;
    \item $(F,F',T)$ is \textit{equivariant}, i.e., $F'(\p s)=F(\p)T(s)$ for every $\p \in \Phi$, $s\in S$.
\end{itemize}
The triple $(F,F',T)$ is called a {\bf perception map} or a {\bf Partial Group Equivariant Operator (P-GEO)} from $(\Phi,{\Phi'},S)$ to $(\Psi,\Psi',Q)$. 
\end{defn}

In Remark \ref{rem_id} we observed that $\text{id}_X\in\Aut_{\Phi,\Phi'}(X)$ if and only if $\Phi\subseteq\Phi'$. Then we can consider a perception triple $(\Phi,\Phi',S)$ with $\Phi\subseteq\Phi'$ and $\text{id}_X\in S\subseteq\Aut_{\Phi,\Phi'}(X)$. Now we will show how a P-GEO from this perception triple behaves.

\begin{lem}\label{lem_P-GEO}
    Consider two perception triples $(\Phi,\Phi',S)$ and $(\Psi,\Psi',Q)$ with domains $X$ and $Y$, respectively, and with $\mathrm{id}_X\in S\subseteq\Aut_{\Phi,\Phi'}(X)$. Let $(F,F',T)$ be a P-GEO from $(\Phi,\Phi',S)$ to $(\Psi,\Psi',Q)$. Then $\Psi\subseteq\Psi'$ and $\mathrm{id}_Y\in Q\subseteq\Aut_{\Psi,\Psi'}(Y)$.
\end{lem}
\begin{proof}
    Since $(F,F',T)$ is a P-GEO, by definition, we have that, for any $s,t\in S$ such that $st \in S$,
    $T(st)=T(s)T(t)$. Since $\text{id}_X\in S$, then
    \begin{equation*}
        T(\text{id}_X)=T(\text{id}_X \text{id}_X)=T(\text{id}_X)T(\text{id}_X)
    \end{equation*}
and hence $T(\text{id}_X)=\text{id}_Y\in Q\subseteq\Aut_{\Psi,\Psi'}(X)$. Moreover, for Remark \ref{rem_id}, we have that $\Psi\subseteq\Psi'$.
\end{proof}
\begin{prop} \label{prop_coinc_intersezione}
    Consider two perception triples $(\Phi,\Phi',S)$ and $(\Psi,\Psi',Q)$ with domains $X$ and $Y$, respectively, and with $\mathrm{id}_X\in S\subseteq\Aut_{\Phi,\Phi'}(X)$. Let $(F,F',T)$ be a P-GEO from $(\Phi,\Phi',S)$ to $(\Psi,\Psi',Q)$. Then $F'|_\Phi=F$.
\end{prop}
\begin{proof}
     Since $(F,F',T)$ is a P-GEO, it is equivariant and by Lemma \ref{lem_P-GEO} we have that
     \[F'(\p)=F'(\p \text{id}_X)=F(\p)T(\text{id}_X)=F(\p)\text{id}_Y=F(\p)\]
     for every $\p\in\Phi$.
\end{proof}

\begin{defn}
Assume that $(\Phi,{\Phi'},S)$ and $(\Psi,\Psi',Q)$ are perception triples. If $(F,F',T)$ is a perception map from $(\Phi,{\Phi'},S)$ to $(\Psi ,\Psi',Q)$ and $F$, $F'$ are non-expansive , i.e.,
\begin{align*}
    \|F(\p_1)-F(\p_2)\|_\infty&\le\|\p_1-\p_2\|_\infty,\\
    \quad \|F'(\p'_1)-F'(\p'_2)\|_\infty&\le\|\p'_1-\p'_2\|_\infty
\end{align*}
for every $\p_1,\p_2 \in \Phi$, $\p'_1,\p'_2\in\Phi'$, then $(F,F',T)$ is called a {\bf Partial Group Equivariant Non-Expansive Operator (P-GENEO)}.
\end{defn}
In other words, a P-GENEO is a triple $(F,F',T)$ such that $F,F'$ are non-expansive and the following diagram commutes for every $s \in S$
\begin{center}
    \begin{tikzcd}
\Phi \arrow[r, "\mathcal{R}_s"] \arrow[d, swap, "F"]
& {\Phi'} \arrow[d, "F'"] \\
\Psi \arrow[r, "\mathcal{R}_{T(s)}"]
& \Psi'
\end{tikzcd}
\end{center}
\begin{rem}
    We can observe that a GENEO (see \cite{bergomi2019towards}) can be represented as a special case of P-GENEO, considering two perception triples $(\Phi,\Phi',S)$, $(\Psi,\Psi',Q)$ such that
     $\Phi=\Phi'$, $\Psi=\Psi'$, and
    the subsets containing the invariant transformations $S$ and $Q$ are groups (and then the map $T\colon S\to Q$ is a homomorphism).
In this setting, a P-GENEO $(F,F',T)$ is a triple where the operators $F$, $F'$ are equal to each other (because of Proposition \ref{prop_coinc_intersezione}) and the map $T$ is a homomorphism. Hence, instead of the triple, we can simply write the pair $(F,T)$, that is a GENEO.
\end{rem}
Considering two perception triples, we typically want to study the space of all P-GENEOs between them with the map $T$ fixed. Therefore, when the map $T$ is fixed and specified, we will simply consider pairs of operators $(F,F')$ instead of triples $(F,F',T)$, and we say that $(F,F')$ is a P-GENEO \textit{associated with}   or \textit{with respect to} the map $T$. Moreover, in this case we indicate the property of equivariance of the triple $(F,F',T)$ writing that the pair $(F,F')$ is $T$-\textit{equivariant}.


\begin{ex}\label{exsquare}
    Let $X=\R^2$. Take a real number $\ell>0$. In $X$ consider the square $Q_1:=[0,\ell]\times[0,\ell]$, and its translation $s_a$ of a vector $a = (a_1,a_2)\in\R^2$ $Q_1':=[a_1,\ell+a_1]\times[a_2,\ell+a_2]$. Analogously, let us consider a real number $0<\eps<\ell$ and two squares inside $Q_1$ and $Q_1'$,  $Q_2:=[\eps,\ell-\eps]\times[\eps,\ell-\eps]$ and   $Q'_2:=[a_1+\eps,\ell+a_1-\eps]\times[a_2+\eps,\ell+a_2-\eps]$, as in Figure \ref{fig:p-geneo}.\\
    \begin{figure}[htp]
        \centering
        \includegraphics[width=11cm]{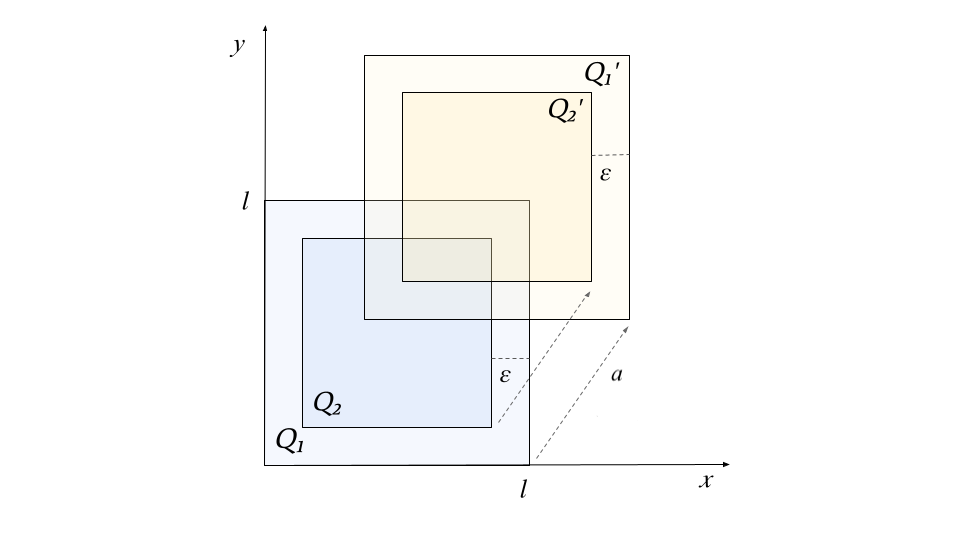}
        \caption{Squares used in Example \ref{exsquare}.}
        \label{fig:p-geneo}
    \end{figure}

Consider the following function spaces in $\R_b^X$:
    \begin{align*}
       &\Phi:=\{\p\colon X\to\R \ | \ \text{supp}(\p)\subseteq Q_1\} \\
       &\Phi':=\{\p'\colon X\to\R \ | \ \text{supp}(\p')\subseteq Q'_1\} \\
       &\Psi:=\{\psi\colon X\to\R \ | \ \text{supp}(\psi)\subseteq Q_2\}\\
       &\Psi':=\{\psi'\colon X\to\R \ | \ \text{supp}(\psi')\subseteq Q_2'\}.
    \end{align*}
Let $S:=\{s_a^{-1}\}$, where $s$ is the translation by the vector $a = (a_1,a_2)$. The triples $(\Phi,\Phi',S)$ and $(\Psi,\Psi',S)$ are perception triples.
This example could model the translation of two nested grey-scale images. We want to build now an operator between these images in order to obtain a transformation that commutes with the selected translation.
We can consider the triple of functions $(F,F',T)$ defined as follows. $F\colon\Phi\to\Psi$ is the operator that maintains the output of functions in $\Phi$ at points of $Q_2$ and set them to zero outside it;  analogously $F'\colon\Phi'\to\Psi'$ is the operator that maintains the output of functions in $\Phi'$ at points of $Q_2'$ and set them to zero outside it; and $T=\mathrm{id}_S$. Therefore, the triple $(F,F',T)$ is a P-GENEO from $(\Phi,\Phi',S)$ to $(\Psi,\Psi',S)$. It turns out that the maps are non-expansive and the equivariance  holds:
\[F'(\p s_a^{-1})=F(\p)T(s_a^{-1})=F(\p)s_a^{-1}\]
for any $\p\in\Phi$. From the point of view of application, we are considering two square images and their translations and we apply an operator that ‘cuts' the images, taking into account only the part of the image that interests the observer.
This example justifies the definition of P-GENEO as a triple of operators $(F,F',T)$, without requiring $F$ and $F'$ to be equal in the possibly non-empty intersection of their domains. In fact, if $\p$ is a function contained in $\Phi\cap\Phi'$, its image via $F$ and $F'$ may be different. 
\end{ex}
\subsection{Methods to construct P-GENEOs} \label{chap3}
Starting from a finite number of P-GENEOs, we will illustrate some methods to construct  new P-GENEOs.
First of all, the composition of two P-GENEOs is still a P-GENEO.
\begin{prop}
    Given two composable P-GENEOs, $(F_1,F_1',T_1)\colon(\Phi,\Phi',S)\to(\Psi,\Psi',Q)$ and $(F_2,F_2',T_2)\colon(\Psi,\Psi',Q)\to(\Omega,\Omega',K)$,
    their composition defined as 
    \[(F,F',T):=(F_2\circ F_1,F_2'\circ F_1',T_2\circ T_1)\colon(\Phi,{\Phi'},S)\to(\Omega,\Omega',K)\]
    is a P-GENEO.
\end{prop}
\begin{proof}
    First, one could easily check that the map $T=T_2\circ T_1$ respects the second and the third property of~\ref{wfbjbfrbhjf}. 
    Therefore, it remains to verify that $F(\Phi)\subseteq\Omega$, $F'(\Phi')\subseteq\Omega'$ and that the properties of equivariance and non-expansiveness are maintained.
    \begin{enumerate}
        \item Since $F_1(\Phi)\subseteq \Psi$ and $F_2(\Psi)\subseteq\Omega$, then we have that $F(\Phi)=(F_2\circ F_1)(\Phi)=F_2(F_1(\Phi))\subseteq F_2(\Psi)\subseteq\Omega$.
    Analogously, $F'(\Phi')\subseteq \Omega'$.
        \item Since $(F_1,F_1',T_1)$ and $(F_2,F_2',T_2)$ are equivariant, then $(F,F',T)$ is equivariant. Indeed, for every $\p\in\Phi$ we have that
        \begin{align*}
            F'(\p s)&=(F_2'\circ F_1')(\p s)=F_2'(F_1'(\p s))\\
            &=F_2'(F_1(\p)T_1(s))=F_2(F_1(\p))T_2(T_1(s))\\
            &=(F_2 \circ F_1)(\p)(T_2 \circ T_1)(s)=F(\p)T(s).
        \end{align*}
        \item Since $F_1$ and $F_2$ are non-expansive, then $F$ is non-expansive; indeed for every $\p_1,\p_2\in\Phi$ we have that
        \begin{align*}
           \|F(\p_1)-F(\p_2)\|_\infty&=\|(F_2\circ F_1)(\p_1)-(F_2\circ F_1)(\p_2)\|_\infty\\
           &=\|F_2(F_1(\p_1))-F_2(F_1(\p_2))\|_\infty\\
           &\le \|F_1(\p_1)-F_1(\p_2)\|_\infty\\
           &\le \|\p_1-\p_2\|_\infty.
        \end{align*} 
        Analogously, $F'$ is non-expansive.
    \end{enumerate}
    
\end{proof}
Given a finite number of P-GENEOs with respect to the same map $T$, we illustrate a general method to construct a new operator as a combination of them. Given two sets $X$ and $Y$, consider a finite set $\{H_1,\dots, H_n\}$ of functions from $\Omega\subseteq\R_b^X$ to $\R_b^Y$ and a map $\mathcal{L}\colon\R^n\to\R$, where $\R^n$ is endowed with the norm $\|(x_1,\dots,x_n)\|_\infty:=\max_{1\le i\le n}|x_i|$. We define $\mathcal{L}^*(H_1,\dots,H_n)\colon \Omega\to \R_b^Y$ as 
\[\mathcal{L}^*(H_1,\dots,H_n)(\omega):=[\mathcal{L}(H_1(\omega),\dots,H_n(\omega))],\]
for any $\omega\in\Omega$, where $[\mathcal{L}(H_1(\omega),\dots,H_n(\omega))]\colon Y \to \R$ is defined by setting
\[[\mathcal{L}(H_1(\omega),\dots,H_n(\omega))](y):=\mathcal{L}(H_1(\omega)(y),\dots,H_n(\omega)(y))\]
for any $y\in Y$.
Now, consider two perception triples $(\Phi,{\Phi'},S)$ and $(\Psi,\Psi',Q)$ with domains $X$ and $Y$, respectively, and a finite set of P-GENEOs $(F_1,F_1'),\dots(F_n,F_n')$ between them associated with the map $T\colon S\to Q$.
We can consider the functions $\mathcal{L}^*(F_1,\dots,F_n)\colon \Phi\to \R_b^Y$ and $\mathcal{L}^*(F_1',\dots,F_n')\colon \Phi'\to \R_b^Y$, defined as before, and state the following result. 

\begin{prop} \label{prop_build}
    Assume that $\mathcal{L}\colon\R^n\to\R$ is non-expansive. If 
    $\mathcal{L}^*(F_1,\dots,F_n)(\Phi)\subseteq\Psi$ and $\mathcal{L}^*(F_1',\dots,F_n')(\Phi')\subseteq\Psi'$,
    then $(\mathcal{L}^*(F_1,\dots,F_n),\mathcal{L}^*(F_1',\dots,F_n'))$ is a P-GENEO from $(\Phi,{\Phi'},S)$ to $(\Psi,\Psi',Q)$ with respect to $T$.
 \end{prop}
\begin{proof}
    By hypothesis, $\mathcal{L}^*(F_1,\dots,F_n)(\Phi)\subseteq\Psi$ and $\mathcal{L}^*(F_1',\dots,F_n')(\Phi')\subseteq\Psi'$, so we just need to verify the properties of equivariance and non-expansiveness.
    \begin{enumerate}
        \item Since $(F_1,F_1'),\dots,(F_n,F_n')$ are $T$-equivariant, then for any $\p\in\Phi$ and any $s\in S$ we have that:
        \begin{align*}
            \mathcal{L}^*(F_1',\dots,F_n')(\p s)&=[\mathcal{L}(F_1'(\p s),\dots,F_n'(\p s))]\\
            &=[\mathcal{L}(F_1(\p)T(s),\dots,F_n(\p)T(s))]\\
            &=[\mathcal{L}(F_1(\p),\dots,F_n(\p))]T(s)\\
            &= \mathcal{L}^*(F_1,\dots,F_n)(\p)T(s).
        \end{align*}
        Therefore $(\mathcal{L}^*(F_1,\dots,F_n),\mathcal{L}^*(F_1',\dots,F_n'))$ is $T$-equivariant.
        \item Since $F_1,\dots,F_n$ and $\mathcal{L}$ are non-expansive, then for any $\p_1,\p_2\in\Phi$ we have that:
        \begin{align*}
        &\|\mathcal{L}^*(F_1,\dots,F_n)(\p_1)-\mathcal{L}^*(F_1,\dots,F_n)(\p_2)\|_\infty\\
            &=\max_{y\in Y}\lvert[\mathcal{L}(F_1(\p_1),\dots,F_n(\p_1))](y)-[\mathcal{L}(F_1(\p_2),\dots,F_n(\p_2))](y)\rvert \\
            &=\max_{y\in Y}|\mathcal{L}(F_1(\p_1)(y),\dots,F_n(\p_1)(y))-\mathcal{L}(F_1(\p_2)(y),\dots,F_n(\p_2)(y))| \\
            &\le \max_{y\in Y}\|(F_1(\p_1)(y)-F_1(\p_2)(y),\dots,F_n(\p_1)(y)-F_n(\p_2)(y))\|_\infty\\
            &=\max_{y\in Y}\max_{1 \le i \le n}|F_i(\p_1)(y)-F_i(\p_2)(y)| \\
            &= \max_{1 \le i \le n} \|F_i(\p_1)-F_i(\p_2)\|_\infty\\
            &\le \|\p_1-\p_2\|_\infty.
        \end{align*}
        Hence, $\mathcal{L}^*(F_1,\dots,F_n)$ is non-expansive. Analogously, since $F_1',\dots,F_n'$ and $\mathcal{L}$ are non-expansive, then $\mathcal{L}^*(F_1',\dots,F_n')$ is non-expansive.
    \end{enumerate}
    Therefore $(\mathcal{L}^*(F_1,\dots,F_n),\mathcal{L}^*(F_1',\dots,F_n'))$ is a P-GENEO from $(\Phi,{\Phi'},S)$ to $(\Psi,\Psi',Q)$ with respect to $T$.
\end{proof}

\begin{rem}
    The above result describes a general method to build new P-GENEOs, starting from a finite number of known P-GENEOs via non-expansive maps. Some examples of such non-expansive maps are the maximum function, the power mean and the convex combination (for further details, see \cite{FrQu17,amsdottorato9770,Qu21}).
\end{rem}
\subsection{Compactness and convexity of the space of P-GENEOs}  
Given two perception triples, under some assumptions on the data sets, it is possible to show two useful features in applications: compactness and convexity.
These two properties guarantee, on the one hand, that the space of P-GENEOs can be approximated by a finite subset of them, and, on the other, that a convex combination of P-GENEOs is again a P-GENEO.

First, we define a metric on the space of P-GENEOs. Let $X,Y$ be sets and consider two sets $\Omega\subseteq\R_b^X,\Delta\subseteq\R_b^Y$, we can define the distance
\[D_{\textbf{NE}}^\Omega(F_1,F_2):=\sup_{\omega\in \Omega}\|F_1(\omega)-F_2(\omega)\|_\infty\]
for every $F_1,F_2\in\textbf{NE}(\Omega,\Delta)$. \\
The metric $D_{\text{P-GENEO}}$ on the space $\mathcal{F}_T^{all}$ of all the P-GENEOs between the perception triples $(\Phi,{\Phi'},S)$ and $(\Psi,\Psi',Q)$ associated with the map $T$ is defined as
\begin{align*}
    D_{\text{P-GENEO}}((F_1,F_1'),(F_2,F_2')):=\max\{D_{\textbf{NE}}^\Phi(F_1,F_2), D_{\textbf{NE}}^{\Phi'}(F'_1,F'_2)\}\\
    \\
    =\max\{\sup_{\p\in \Phi}\|F_1(\p)-F_2(\p)\|_\infty,\sup_{\p'\in \Phi'}\|F_1'(\p')-F_2'(\p')\|_\infty\}
\end{align*}
for every $(F_1,F_1'),(F_2,F_2')\in\mathcal{F}_T^{all}$. 
\subsubsection{Compactness}
Before proceeding, we need to prove that the following result holds:
\begin{lem}\label{cor_ascoli_arzela}
If $(P,d_P),(Q,d_Q)$ are compact metric spaces, then $\textbf{NE}(P,Q)$ is compact.
\end{lem}

\begin{proof}
    Theorem 5 in \cite{li2012improvement} implies that $\textbf{NE}(P,Q)$ is relatively compact, since it is a equicontinuous space of maps.
    Hence, it will suffice to show that $\textbf{NE}(P,Q)$ is closed. Considering a sequence $(F_i)_{i\in\mathbb{N}}$ in $\textbf{NE}(P,Q)$ such that $\lim_{i\to\infty}F_i=F$, we have that
    \begin{align*}
        d_Q(F(p_1),F(p_2))=\lim_{i\to\infty}d_Q(F_i(p_1), F_i(p_2))\le d_P(p_1,p_2)
    \end{align*}
    for every $p_1,p_2\in P$. Therefore, $F\in\textbf{NE}(P,Q)$. It follows that $\textbf{NE}(P,Q)$ is closed.
\end{proof}


Consider two perception triples $(\Phi,\Phi',S)$ and 
$(\Psi,\Psi',Q)$, with domains $X$ and $Y$, respectively, and the space $\mathcal{F}_T^{all}$ of P-GENEOs between them associated with the map $T\colon S\to Q$. The following result holds:
\begin{thm} \label{thm_compact_P-GENEO}
If $\Phi,\Phi',\Psi$ and $\Psi'$ are compact, then $\mathcal{F}_T^{all}$ is compact with respect to the metric $D_{\mathrm{P-GENEO}}$.
\end{thm}
\begin{proof}
    By definition, $\mathcal{F}_T^{all}\subseteq\textbf{NE}(\Phi,\Psi)\times \textbf{NE}(\Phi',\Psi')$. 
    Since  $\Phi,\Phi',\Psi$ and $\Psi'$ are compact, for Lemma \ref{cor_ascoli_arzela} the spaces $\textbf{NE}(\Phi,\Psi)$ and $\textbf{NE}(\Phi',\Psi')$ are also compact, and then, by Tychonoff's Theorem, the product $\textbf{NE}(\Phi,\Psi)\times \textbf{NE}(\Phi',\Psi')$ is also compact, with respect to the product topology. 
    Hence, to prove our statement it suffices to show that $\mathcal{F}_T^{all}$ is closed. 
    Let us consider a sequence $((F_i,F_i'))_{i\in\mathbb{N}}$ of P-GENEOs, converging to a pair $(F,F') \in \textbf{NE}(\Phi,\Psi)\times \textbf{NE}(\Phi',\Psi')$.
    Since $(F_i,F_i')$ is $T$-equivariant for every $i\in\mathbb{N}$ and the action of $Q$ on $\Psi$ is continuous (see Proposition \ref{azione_continua}), $(F,F')$ belongs to $\mathcal{F}_T^{all}$. Indeed, we have that
    \begin{equation*}
        F'(\p s)=\lim_{i\to\infty}F'_i(\p s)=\lim_{i\to\infty}F_i(\p)T(s)=F(\p)T(s)
    \end{equation*}
    for every $s\in S$ and every $\p \in \Phi$. 
    Hence, $\mathcal{F}_T^{all}$ is a closed subset of a compact set and then it is also compact.
\end{proof}

\subsubsection{Convexity}
Assume that $\Psi,\Psi'$ are convex. Let $(F_1,F_1'),\dots,(F_n,F_n')\in\mathcal{F}_T^{all}$ and consider an $n$-tuple $(a_1,\dots,a_n)\in\R^n$ with $a_i\geq0$ for every $i\in\{1,\dots, n\}$ and $\sum_{i=1}^n a_i= 1$. We can define two operators $F_\Sigma\colon\Phi\to\Psi$ and  $F_\Sigma'\colon\Phi'\to\Psi'$ as
\begin{equation*}
    F_\Sigma(\p):=\sum_{i=1}^n a_iF_i(\p), \ \text{and} \ F_\Sigma'(\p'):=\sum_{i=1}^n a_iF_i'(\p')
\end{equation*}
for every $\p \in \Phi,\p'\in\Phi'$. We notice that the convexity of $\Psi$ and $\Psi'$ guarantees that $F_\Sigma$ and $F_\Sigma'$ are well defined.

\begin{prop} \label{prop_convex}
    $(F_\Sigma,F_\Sigma')$ belongs to $\mathcal{F}_T^{all}$.
\end{prop}
\begin{proof}
    By hypothesis, for every $i\in\{1,\dots,n\}$ $(F_i,F_i')$ is a perception map, and then:
    \begin{align*}
        F_\Sigma'(\p s)=\sum_{i=1}^n a_iF'_i(\p s)&=\sum_{i=1}^n a_i(F_i(\p) T(s))\\
        &=\Bigl( \sum_{i=1}^n a_i F_i(\p) \Bigr) T(s)\\
        &=F_\Sigma(\p)T(s)
    \end{align*}
    for every $\p \in\Phi$ and every $s\in S$. Furthermore, since for every $i\in\{1,\dots, n\}$ $F_i(\Phi)\subseteq\Psi$ and $\Psi$ is convex, also $F_\Sigma(\Phi)\subseteq\Psi$. Analogously, the convexity of $\Psi'$ implies that $F_\Sigma'(\Phi')\subseteq\Psi'$. Therefore $(F_\Sigma,F_\Sigma')$ is a P-GEO. It remains to show the non-expansiveness of $F_\Sigma$ and $F_\Sigma'$. Since $F_i$ is non-expansive for any $i$, then for every $\p_1,\p_2\in\Phi$ we have that
    \begin{align*}
        \|F_\Sigma(\p_1)-F_\Sigma(\p_2)\|_\infty&=\left\|\sum_{i=1}^n a_iF_i(\p_1)-\sum_{i=1}^n a_iF_i(\p_2)\right\|_\infty \\
        &=\left\|\sum_{i=1}^n a_i(F_i(\p_1)-F_i(\p_2))\right\|_\infty \\
        &\le\sum_{i=1}^n |a_i|\left\|F_i(\p_1)-F_i(\p_2)\right\|_\infty\\
        &\le\sum_{i=1}^n |a_i|\|\p_1-\p_2\|_\infty=\|\p_1-\p_2\|_\infty.
    \end{align*}
    Analogously, since every $F_i'$ is non-expansive, for every $\p_1',\p_2'\in\Phi'$ we have that
    \[\|F_\Sigma'(\p'_1)-F_\Sigma'(\p'_2)\|_\infty\le\sum_{i=1}^n |a_i|\|\p_1'-\p_2'\|_\infty=\|\p_1'-\p_2'\|_\infty.\]
    Therefore, we have proven that $(F_\Sigma,F_\Sigma')$ is a P-GEO with $F_\Sigma$ and $F_\Sigma'$ non-expansive. Hence it is a P-GENEO.
\end{proof}
Then, the following result holds:
\begin{cor}
    If $\Psi,\Psi'$ are convex, then the set $\mathcal{F}_T^{all}$ is convex.
\end{cor}
\begin{proof}
    It is sufficient to apply Proposition \ref{prop_convex} for $n=2$, by setting $a_1=t$, $a_2=1-t$ for $0\le t\le 1$.
\end{proof}

\section{Conclusions}
In this article we proposed a generalization of some known results in the theory of GENEOs to a new mathematical framework, where the collection of all symmetries is represented by a subset of a group of transformations.
We introduced P-GENEOs and showed that they are a generalisation of GENEOs.
We defined pseudo-metrics on the space of measurements and on the space of P-GENEOs and studied their induced topological structures.
Under the assumption that the function spaces are compact and convex, we showed compactness and convexity of the space of P-GENEOs. 
In particular, compactness guarantees that any operator can be approximated by a finite number of operators belonging to the same space, while convexity allows us to build new P-GENEOs by taking convex combinations of P-GENEOs. 
Compactness and convexity together ensure that every strictly convex loss function on the space of P-GENEOs admits a unique global minimum.
Given a collection of P-GENEOs, we presented a general method to construct new P-GENEOs as combinations of the initial ones.


\bibliographystyle{plain}
\bibliography{bib}

\end{document}